\documentclass[11pt]{article}

\usepackage{microtype} 
\usepackage{booktabs}  
\usepackage{url}  
\usepackage{csquotes}
\usepackage{amsmath}
\usepackage{amsthm}

\usepackage{graphicx}
\usepackage{subcaption}

\usepackage{algorithm}
\usepackage{algpseudocode}
\usepackage{algcompatible}
\newcommand{\RETURN}{\STATE \textbf{return} }

\usepackage[inline]{enumitem}

\usepackage{multirow}

\usepackage[final]{automl}
\makeatletter
\renewcommand{\c@nferencestring}{}
\makeatother

\usepackage{natbib}
\usepackage{wrapfig}
\usepackage{filecontents}

\newcommand{\NAME}{HAPEns}
\title{\NAME: Hardware-Aware Post-Hoc Ensembling \\ for Tabular Data}

\author[1,2]{\nameemail{Jannis Maier}{jannis.maier@helmholtz-berlin.de}}
\author[3,4]{\nameemail{Lennart Purucker}{purucker@cs.uni-freiburg.de}}

\makeatletter
\renewcommand{\AB@affilsepx}{; \protect\Affilfont}
\makeatother
\affil[1]{Helmholtz-Zentrum Berlin}
\affil[2]{TU Dortmund}
\affil[3]{Prior Labs} 
\affil[4]{University of Freiburg}

\hypersetup{%
  pdfauthor={AutoML}, 
  pdftitle={Documentation for the automl Package},
  pdfsubject={Documentation for automl Package},
  pdfkeywords={AutoML, LaTeX, style}
}

\begin{document}

\maketitle

\begin{abstract}
Ensembling is commonly used in machine learning on tabular data to boost predictive performance and robustness, but larger ensembles often lead to increased hardware demand. 
We introduce \NAME, a post-hoc ensembling method that explicitly balances accuracy against hardware efficiency. 
Inspired by multi-objective and quality diversity optimization, \NAME\ constructs a diverse set of ensembles along the Pareto front of predictive performance and resource usage.
Existing hardware‑aware post‑hoc ensembling baselines are not available, highlighting the novelty of our approach. 
Experiments on 83 tabular classification datasets show that \NAME\ significantly outperforms baselines, finding superior trade-offs for ensemble performance and deployment cost. 
Ablation studies also reveal that memory usage is a particularly effective objective metric. 
Further, we show that even a greedy ensembling algorithm can be significantly improved in this task with static multi-objective weighting.
\end{abstract}


\begin{figure}[h!]
    \centering
    \includegraphics[width=0.7\linewidth]{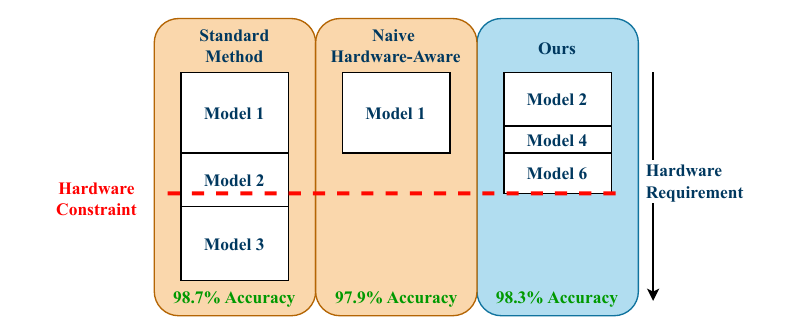}
    \caption{\textbf{Illustration of three ensemble selection strategies:} a standard method ignoring hardware constraints, a naive hardware-aware variant that sacrifices accuracy, and an advanced hardware-aware method that balances accuracy and efficiency. Box size reflects model resource usage; the red dashed line indicates the hardware resource constraint.}
    \label{fig:enter-label}
\end{figure}

\section{Introduction}
%
Ensembling is a central technique in machine learning, used to improve predictive performance, stability, and robustness across a wide range of applications. 
From boosting and bagging in classical supervised learning to stacking in modern deep learning workflows, ensembles are frequently adopted to combine the strengths of diverse models. 
In many practical scenarios, models produced during training or exploratory analysis are later combined into ensembles in a post-hoc fashion to substantially improve performance~\citep{erickson_tabarena_2025, arango_regularized_2025}. 
This workflow has been further popularized by automated machine learning (AutoML) systems for tabular data \citep{purucker_assembled-openml_2023, he_automl_2021, erickson_autogluon-tabular_2020}, where greedy ensemble selection (GES) by \cite{caruana_ensemble_2004} has emerged as a widely used method to automatically build strong ensembles from model libraries. 

While post-hoc ensembling generally improves predictive performance, larger ensembles increase hardware demands at inference time. 
Each additional model increases prediction latency and resource consumption, inducing higher costs. While this matters greatly in production settings, standard post-hoc ensembling methods ignore it. 
As machine learning is increasingly deployed in environments with tight resource constraints, the gap between high predictive accuracy and hardware feasibility has become more pronounced. \\
We address this challenge by introducing \NAME, a post-hoc ensembling method that explicitly balances predictive performance against hardware costs. 
It improves on existing baselines by constructing Pareto fronts for ensembles that better balance competing objectives.
Thus, practitioners can select models that satisfy their performance and deployment requirements.
Drawing inspiration from multi-objective optimization~\citep{gunantara_review_2018} and quality diversity optimization~\citep{pugh_quality_2016}, \NAME\ maintains a diverse population of ensembles that vary in both hardware cost and predictive behavior, while optimizing for predictive performance. 
The result is a set of candidate ensembles that offer trade-offs between both objectives. 

To evaluate \NAME, we conducted experiments across 83 tabular classification datasets of varying sizes and complexities.
All cost metrics are concrete measurements taken on the same system, held fixed across methods for fair comparison.
We compare ensembles constructed by our method with those selected by baselines, including GES and a novel multi-objective baseline. 
Our findings reveal that optimizing for memory footprint is a particularly effective metric for deployment cost and that our method significantly outperforms competitors in balancing hardware costs and predictive performance.
To our knowledge, this is the first systematic study of hardware-aware post-hoc ensemble selection. 
Prior hardware-aware work focuses on model search or NAS, rather than on ensemble construction from fixed model libraries.

\paragraph{Our Contributions.} In this work, we: 
\begin{enumerate*}[label=(\roman*)] 
\item Propose a novel post-hoc ensembling algorithm that explicitly incorporates hardware cost into the selection process; 
\item Demonstrate through extensive benchmarking that our method achieves superior accuracy–cost trade-offs compared to existing baselines; 
\item Show that memory-awareness yields substantial gains even in inference-time efficiency; 
\item Limitations, including the dependence on a single hardware configuration, are discussed, with directions for extending the method to heterogeneous devices.
\item Ensure reproducibility by open-sourcing all code\footnote{All code used for this publication is available at: https://github.com/Atraxus/HA-ES}, results, and integration with popular ensembling frameworks. 
\end{enumerate*}

\section{Related Work}

\begin{figure}[ht]
    \centering
    \includegraphics[width=0.8\linewidth]{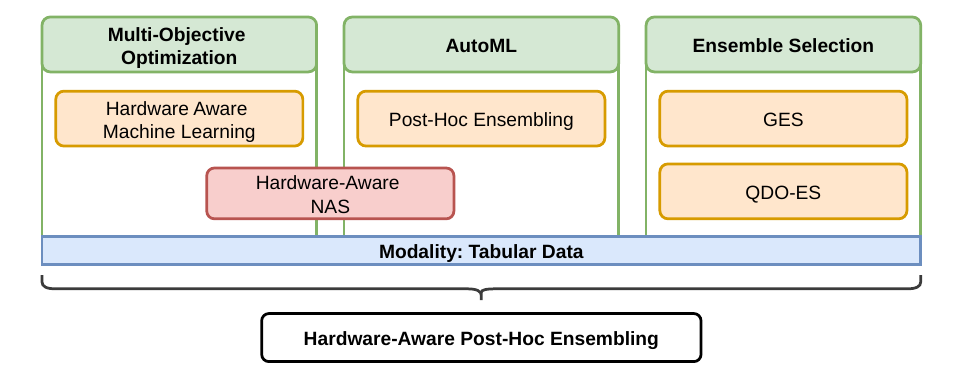}
    \caption{Overview of the main research areas. HW-NAS (red) is shown as a parallel area, while the others (orange) directly influence \NAME. This work focuses solely on tabular data (blue).}
    \label{fig:rel-work}
\end{figure}

Ensembling\textemdash combining multiple pre-trained models\textemdash is an effective approach to improve predictive performance and robustness. 
Common strategies include bagging, stacking, and ensemble selection (ES).
Bagging and stacking are typically integrated into the training process, whereas ES is applied post hoc, that is, after model training has completed. 
ES is thus called post hoc ensembling.
Figure~\ref{fig:rel-work} provides an overview of the research fields discussed in the following paragraphs.

ES, as introduced by \cite{caruana_ensemble_2004}, is a forward selection algorithm that greedily constructs an ensemble by iteratively adding the model that most improves the ensemble's predictive performance. 
The resulting ensemble is defined by a weight vector derived from the selected models in this superset. 
We adopt a broader interpretation of ensemble selection: Any algorithm that produces such a weight vector from a pool of trained models qualifies as ES. To distinguish the classical algorithm of Caruana et al., we refer to it as greedy ensemble selection (GES).

Post-hoc ensembling is a widely adopted component in automated machine learning (AutoML) systems, particularly for tabular data~\citep{erickson_autogluon-tabular_2020, feurer_efficient_2015, purucker_assembled-openml_2023}. 
It enables the reuse of models generated during training without retraining, making it a computationally attractive final optimization step. 
Although ensemble selection can theoretically be used during training, we reserve the term ES for its post-hoc usage in this work. 
The term blending also appears in this context, but it specifically refers to ensemble selection applied to a holdout validation set distinct from the training data.

Recent years have seen the integration of multi-objective optimization (MOO) into various stages of the machine learning and AutoML pipelines, including neural architecture search (NAS)~\citep{benmeziane_comprehensive_2021, benmeziane_hardware-aware_2021}. 
These methods optimize for trade-offs such as accuracy versus latency, energy consumption, or memory usage. 
However, the use of MOO techniques in post-hoc ensemble selection remains largely unexplored. 
\citet{shen_divbo_2022} introduced DivBO, a diversity-aware Bayesian optimization framework that incorporates ensemble selection during candidate evaluation to promote both accuracy and diversity. 
Although their approach targets the model search stage rather than post-hoc optimization, it highlights the potential of multi-objective formulations to improve ensemble composition.
Nevertheless, to the best of our knowledge, no prior work systematically investigates the construction of Pareto-optimal ensembles that explicitly account for hardware constraints such as inference time or memory usage.

Modern implementations of GES\textemdash still the de facto standard in AutoML frameworks such as Auto-sklearn~\citep{feurer_auto-sklearn_2022} and AutoGluon~\citep{erickson_autogluon-tabular_2020}\textemdash typically optimize only for predictive performance and remain agnostic to deployment cost. 
Consequently, they may produce ensembles that are unnecessarily large or infeasible for deployment due to hardware requirements. 

Our work addresses this identified gap by introducing a hardware-aware approach to ES, explicitly targeting the trade-off between accuracy and resource usage. 
QDO-ES as developed by Purucker et al.~\citep{purucker_qdo-es_2023} inspired \NAME\ and the inclusion cost metrics during ensemble construction. 
In doing so, we extend the utility of ES beyond predictive performance to deployment and real-world use.

\section{Method}
One of the last steps in the ML pipeline is model generation, where human experts or AutoML systems explore and evaluate various configurations. This process yields a set of candidate models, which is typically followed by selecting the best single model for deployment. Post‑hoc ensembling, instead, aims to improve prediction quality by combining multiple models from this set. 

Let \(\mathcal{M}=\{M_1,\dots,M_p\}\) be the library of models and let \(c_j\) be the number of times \(M_j\) is selected out of a total of \(T\) picks (with repetition). Define the weight vector $\mathbf{w}$:
\begin{equation}
\mathbf{w}=(w_1,\dots,w_p)^\top
=\frac{1}{T}(c_1,\dots,c_p)^\top,
\quad
w_j=\frac{c_j}{T},\quad\sum_jw_j=1.
\end{equation}
The ensemble predictor for input \(x\) is
$\label{eq:ens-pred}
f_{\mathrm{ens}}(x)
=\sum_{j=1}^p w_j\,f_j(x).
$
This formulation applies broadly: for regression, each \(f_j(x)\) is a scalar prediction; for probabilistic classification, \(f_j(x)\) is a vector of class probabilities, and \(f_{\mathrm{ens}}(x)\) is the averaged probability vector.  

Although the ensemble predictor is ultimately defined by a weight vector, there are multiple ways to construct it. 
A common method is GES, which uses a forward selection strategy to iteratively build the ensemble by greedily adding models that improve performance the most. 
In contrast, our work explores a population-based approach.

\begin{wrapfigure}{r}{0.45\textwidth}
    \centering
    \vspace{-10pt}
    \includegraphics[width=0.48\textwidth]{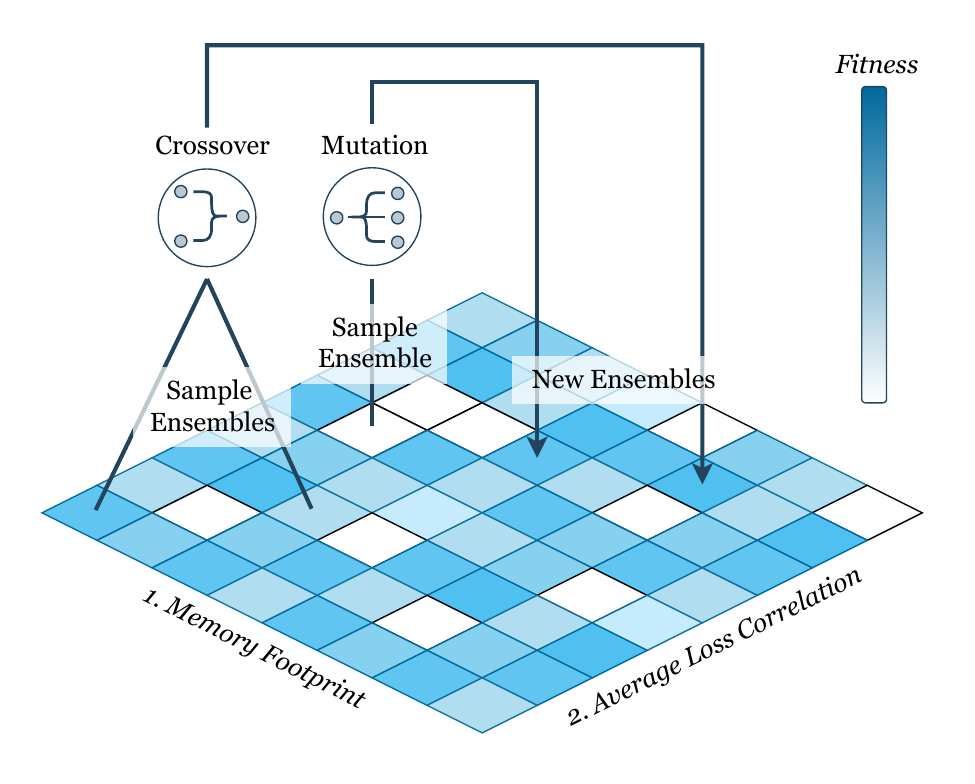}
    \caption{\textbf{The \NAME\ search process.} 
    Ensembles are sampled from bins over memory footprint and average loss correlation, then evolved via crossover and mutation to explore the behavior space.}
    \label{fig:beh-spa-detailed}
    \vspace{-10pt}
\end{wrapfigure}

We begin by sampling an initial population of ensembles across a two-dimensional behavior space (e.g., memory footprint vs. average loss correlation). 
Each ensemble is evaluated and stored in a niche corresponding to its behavior and hardware costs. 
New ensembles are generated by selecting suitable parents from these niches and applying crossover and mutation (see Figure \ref{fig:beh-spa-detailed}). 
This process repeats until convergence or a time/iteration limit is reached, allowing us to explore a wide range of model combinations and discover Pareto‑optimal trade‑offs between prediction quality and deployment cost. 
What follows now are detailed definitions of these concepts, similarly outlined by \cite{purucker_qdo-es_2023}.

\textbf{Behavior Space.}
Each ensemble \(E\) is assigned a two-dimensional descriptor \(b(E) = (\mathrm{ALC}, \mathrm{HW})\), where \(\mathrm{ALC}\) is the mean Pearson correlation among the loss vectors of its constituent models and \(\mathrm{HW}\) is a cost metric aggregated over those models. 
Following prior work~\citep{purucker_qdo-es_2023}, we divide this 2D space into a 7 by 7 grid using a sliding bounding archive~\citep{fontaine_mapping_2019}, creating $49$ bins (niches).\footnote{The 7$\times$7 grid follows the setup of \citet{purucker_qdo-es_2023}, balancing behavior space coverage against niche sample density. Sensitivity to this choice is mitigated by the sliding boundaries archive~\citep{fontaine_mapping_2019}, which adapts niche boundaries to the observed solution distribution; a full sensitivity analysis is left for future work.}
The algorithm allows ensembles to compete only within the same niche. 
This ensures that different regions of the behavior space can retain their best solutions. 
Therefore, a diverse population across two objectives is maintained while optimizing predictive performance.
Here, we found that memory, as a cost metric, produces ensembles that trade off predictive performance and hardware cost best.

\textbf{Fitness.}
Each ensemble \(E\) is scored by a scalar loss \(L(E)\) on cross-validation data. The behavior space is partitioned into fixed niches or bins, and each niche retains the lowest loss ensemble observed.

\textbf{Sampling.}
The parents are selected from the archive using a combined dynamic strategy that balances exploration and exploitation. 
The method alternates between deterministic selection of the best solution and stochastic selection of random solutions, with the selection probability dynamically adjusted every ten iterations based on which approach yields better results. 
Deterministic and tournament-based selection methods are also available as alternatives.

\textbf{Crossover.}
Two parent repetition vectors $r$ and $r'$ are recombined using two-point crossover restricted to the index set $S = \{i \mid r[i] > 0\} \cup \{i \mid r'[i] > 0\}$, i.e., indices nonzero in either parent. 
If $|S| < 3$ or the resulting offspring is all-zero, average crossover is used instead, rounding counts up to the nearest integer to maintain valid repetition counts. 
The procedure is detailed in Algorithm~\ref{alg:crossover_mutation}.

\textbf{Mutation.}
A single element of the child repetition vector $r_{\text{child}}$ is incremented by one, with the index $j$ chosen uniformly at random from the model pool $\mathcal{P}$ of size $p$. 
Rejection sampling avoids re-evaluating previously seen ensembles, allowing up to 50 retries; if all retries fail, an emergency brake increases the increment magnitude or raises the mutation-after-crossover probability to $1.0$ for the current iteration to escape the duplicate region. 
See Algorithm~\ref{alg:crossover_mutation}.

\section{Experimental Setup}
The main objective of this paper is to compare the proposed method and the baselines on how well they can balance predictive performance and hardware costs. In this context, only Pareto optimal ensembles are relevant. In addition, there is no best ensemble because choosing the right trade-off depends on the real world scenario. Therefore, our main focus lies with the Pareto fronts of ensembles generated by each method.

Our proposed method uses memory usage as its hardware-aware behavior metric. We compare it with four baselines:
\begin{enumerate*}[label=(\arabic*)]
    \item \textbf{Single-Best}: A naive baseline that selects the single model with the highest validation performance. Including Single-Best highlights the performance gains achieved through ensembling.
    \item \textbf{GES*}: Our implementation of greedy ensemble selection (GES) enhanced to return the entire sequence of ensembles generated during its run.
    GES* therefore represents the best-case performance of the original widely used GES, providing a strong reference point for assessing improvements.
    \item \textbf{Multi-GES}: Our implementation of novel multi-objective extensions of GES to enable the algorithm to balance predictive performance and inference time using a static weighting scheme; see Appendix~\ref{subsect:enh_ges} for details on our implementation. Multi-GES introduces a naive approach to optimizing multiple objectives in ensemble selection and allows us to assess the benefits of our more flexible formulation.
    \item \textbf{QDO-ES}: The quality-diversity optimization ensemble selection method \citep{purucker_qdo-es_2023}, which optimizes for performance but is not hardware aware. This baseline isolates the effect of hardware awareness by comparing against a method that can already generate various Pareto-optimal ensembles without considering resource costs.
\end{enumerate*}

To assess the quality of the generated Pareto fronts, we rely on two standard multi-objective indicators: inverted generational distance plus (IGD+)~\citep{ishibuchi_modified_2015} and hypervolume (HV)~\citep{zitzler_multiobjective_1999}.
IGD+ quantifies how well a set of solutions approximates a reference front, which in our case is constructed from the Pareto optimal solutions of all the methods under comparison. 
HV measures the portion of the objective space dominated by a set of solutions (see \ref{subsect:exp-setup} for details). 
The set of solutions here is the set of ensembles constructed by one method for a given task and seed.
Both HV and IGD+ are widely used in multi-objective evaluation, and for our experiments we employ the \texttt{pygmo}~\citep{biscani_parallel_2020} implementation. 
We focus primarily on HV in the main analysis because we do not have a true Pareto front for IGD+, and both metrics lead to the same conclusions.

The ROC AUC and cost metrics were normalized per seed and task using min-max normalization over all methods.
This makes the results comparable across experiments even after selecting specific methods per experiment. 
To ensure a comprehensive and reproducible evaluation, we organize our experiments into three groups: (1) Main Results, (2) Details, and (3) Ablation (shown in Table~\ref{tab:exp-overview}).

\begin{figure}[t]
    \centering
    \begin{minipage}[t]{0.45\linewidth}
        \centering
        \includegraphics[width=\linewidth]{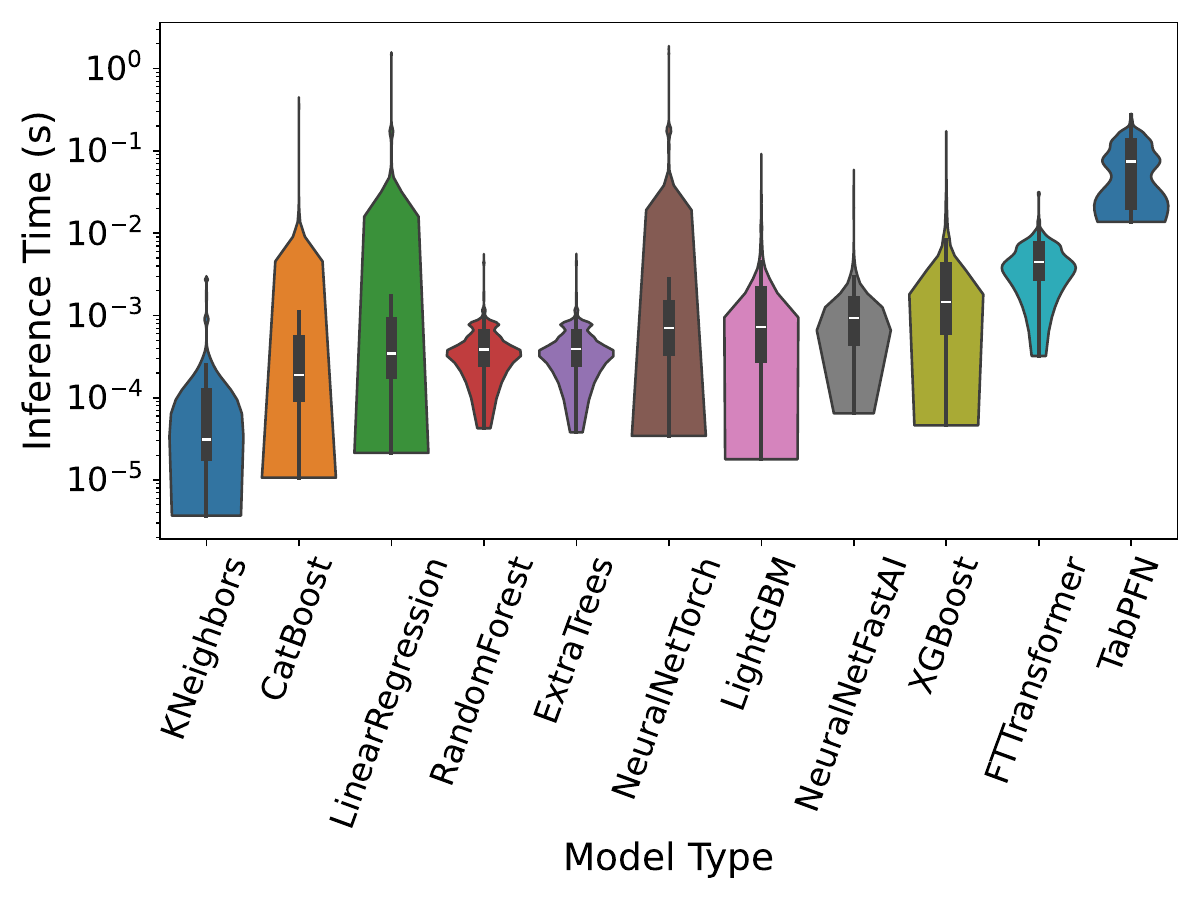}
        \captionsetup{width=\linewidth, format=plain}
        \caption{Comparison of TabRepos model types and their corresponding inference times for varying tasks. 
        }
        \label{fig:tab-basemodels}
    \end{minipage}
    \hfill
    \begin{minipage}[t]{0.53\linewidth}
        \centering
        \includegraphics[width=\linewidth]{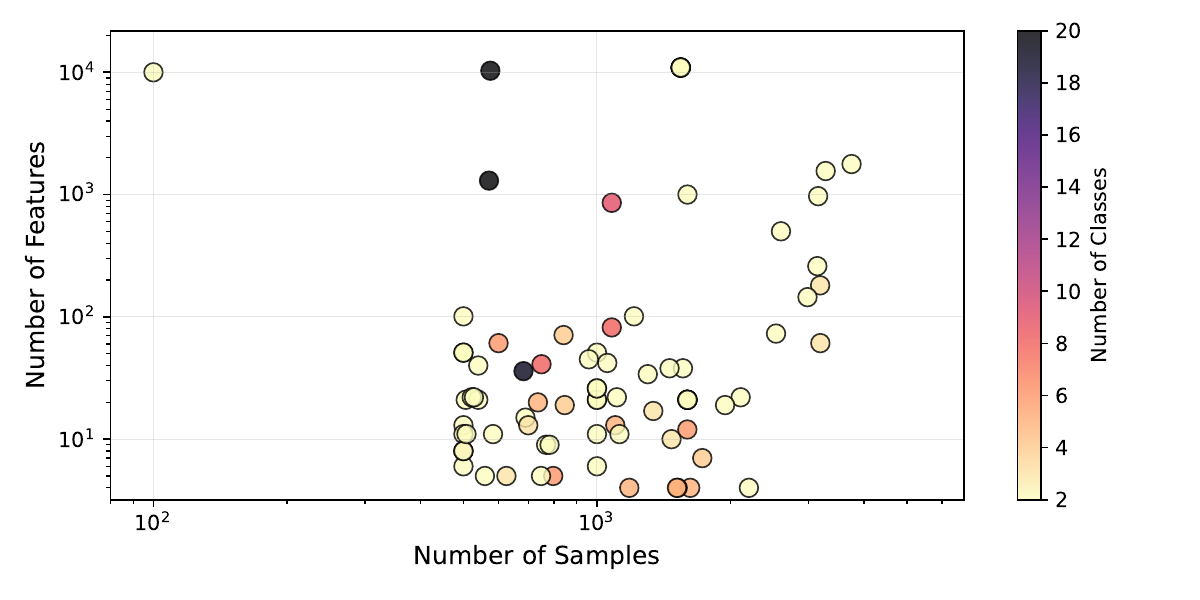}
        \captionsetup{width=\linewidth, format=plain}
        \caption{Scatter plot of datasets over their number of features (y), number of samples (x), and the number of classes (color).}
        \label{fig:tab-datasets}
    \end{minipage}
\end{figure}

\paragraph{Datasets}
We conducted our experiments using TabRepo~\citep{salinas_tabrepo_2023}, which provides precomputed model predictions for 1,530 model configurations across 211 tabular datasets, enabling large-scale, reproducible simulation of post-hoc ensemble selection. 
We used the \texttt{D244\_F3\_C1530\_100} context, a pre-configured evaluation setup that covers 100 of these datasets; after excluding regression tasks, 83 classification datasets remained. 
We aggregated results for 10 seeds to account for run-to-run variance.
Their characteristics are shown in Figure~\ref{fig:tab-datasets}, revealing a wide variety of class, sample, and feature counts.

The available base models are plotted in Figure~\ref{fig:tab-basemodels} with their inference times, illustrating the diversity of the model pool\textemdash from cheap linear and boosting methods to computationally expensive transformers\textemdash providing a realistic and varied set of candidates for ensemble construction. 
Since each component model with non-zero weight must be evaluated independently at inference time, every such model incurs its full hardware cost. 
Therefore, the hardware cost of an ensemble under weight vector $w$ equals the sum of the hardware costs of all models with non-zero weight, i.e., $\sum_{j : w_j \neq 0} h_j$, where $h_j$ denotes the hardware cost of model $M_j$.


\section{Results}
We first present the main results, followed by detailed analyses, and finally ablation studies. 
The central focus is on the ability of each method to balance two objectives: predictive performance and hardware cost. 
In particular, identifying a single strong ensemble may be less effective than discovering several competitive ensembles that trade off these objectives differently. 
In general, \NAME consistently outperforms baselines in both main results and detailed analysis, demonstrating its superior ability to produce competitive ensembles while incorporating hardware awareness.

\subsection*{Main Results (\texttt{EXP1})} 
\label{sub:main-results}

\begin{figure}[t]
    \centering
    \begin{minipage}[t]{0.48\linewidth}
        \centering
        \includegraphics[width=\linewidth]{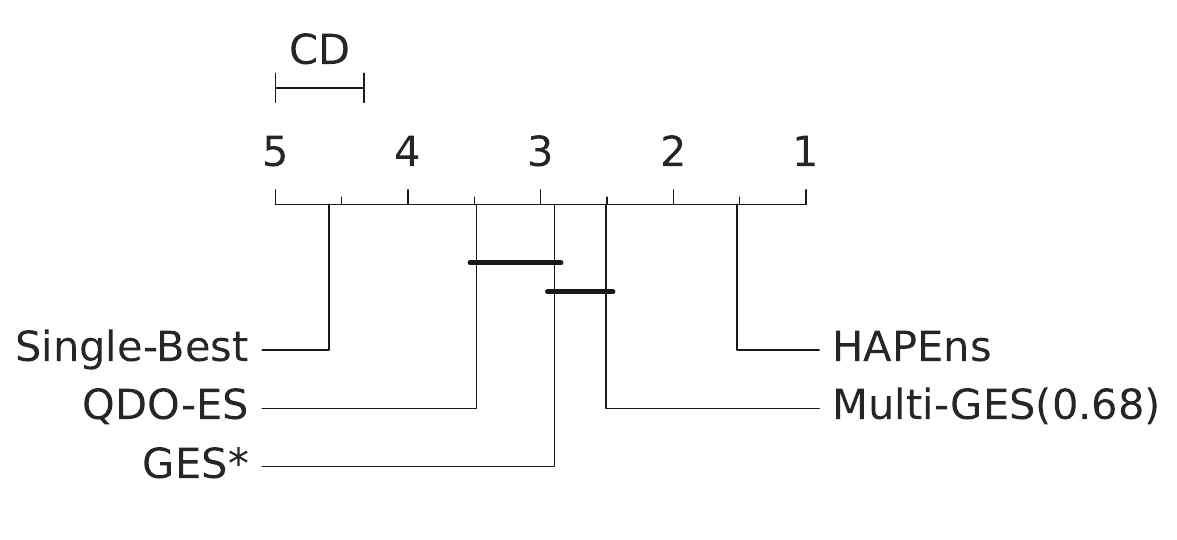}
        \captionsetup{width=\linewidth, format=plain}
        \caption{\NAME\ significantly outperforms the baselines on HV. Single-Best is significantly outperformed by all other methods.}
        \label{fig:main-hv-hwscore}
    \end{minipage}
    \hfill
    \begin{minipage}[t]{0.48\linewidth}
        \centering
        \includegraphics[width=\linewidth]{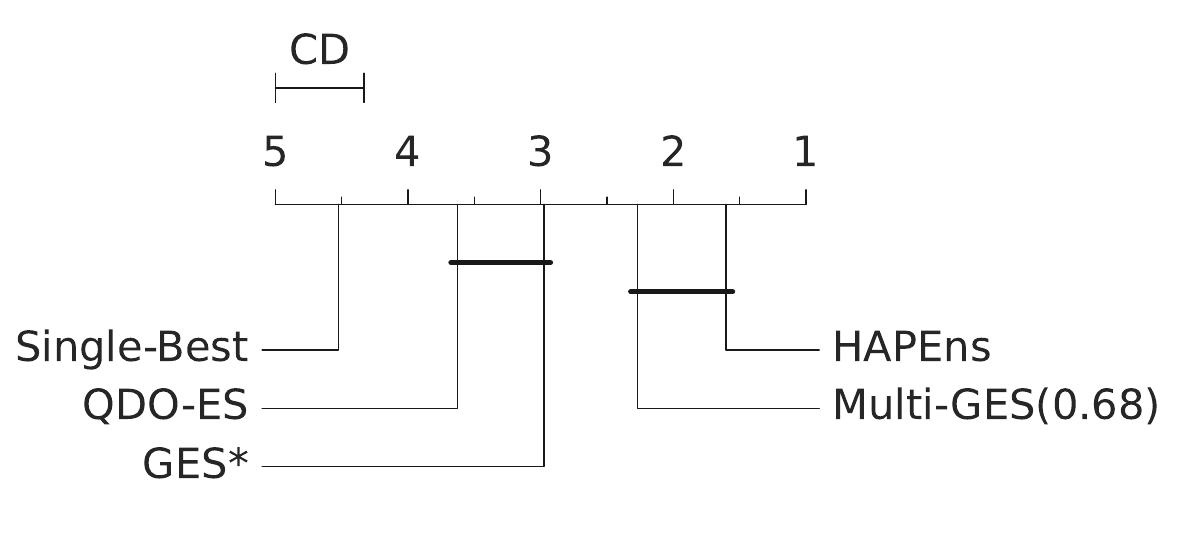}
        \captionsetup{width=\linewidth, format=plain}
        \caption{\NAME\ significantly outperforms the baselines on IGD+. Single-Best is significantly outperformed by all other methods.}
        \label{fig:cdp-igp}
    \end{minipage}
\end{figure}

\paragraph{\texttt{EXP1}}
Figure~\ref{fig:main-hv-hwscore} shows a critical difference (CD) diagram~\citep{demsar_statistical_2006, herbold_autorank_2020} summarizing the average ranks of the methods evaluated based on their HV values.
The HV was calculated from the inverted ROC AUC on the test data and the averaged normalized\footnote{Min-max normalization applied after averaging over folds and seeds} cost metrics (inference time, memory, and disk usage)\textemdash collectively referred to as the \textit{hardware aggregate}.
Therefore, this figure provides an overview for all the datasets, model configurations, and cost metrics we explored in our tests. 
To simplify the presentation and highlight the overall trade-off between predictive performance and hardware costs, we aggregate the three hardware measures into a single score. 
This avoids overemphasizing any single metric, while keeping the focus on the general notion of hardware efficiency. 

In the CD plot, methods connected by a horizontal bar are statistically indistinguishable according to the Nemenyi post-hoc test.
\NAME\ shows significantly superior performance to the baselines, which makes it the best method to balance the trade-off between predictive performance and hardware costs. 
Between the baselines, we do not see significant differences except for the single-best method, which simply picks the best model configuration based on its ROC AUC. 
A single-best model is not well suited for this setting because it cannot capture diverse trade-offs between predictive performance and different hardware costs, which multiple ensembles can exploit more effectively. 
We see slight improvements in GES* over QDO-ES, which can be attributed to the modification of GES to return all intermediate ensembles, which generally leads to a higher number of ensembles produced (see Figure~\ref{fig:pareto-vs-total}). 
This improvement over the standard procedure of returning the final ensemble gives GES a strong edge here.
Multi-GES performs slightly higher, but insignificantly so, by constructing ensembles with reduced hardware costs while keeping their predictive performance comparable to GES*. 
A discussion on GES*'s overfitting problem and the corresponding cost-to-performance trade-off follows in the Multi-GES ablation part of this section.

\subsection*{Details (\texttt{EXP2}, \texttt{EXP3}, \texttt{EXP4})} 
\label{sub:details-results}

\begin{wrapfigure}{r}{0.5\textwidth}
    \centering
    \includegraphics[width=\linewidth]{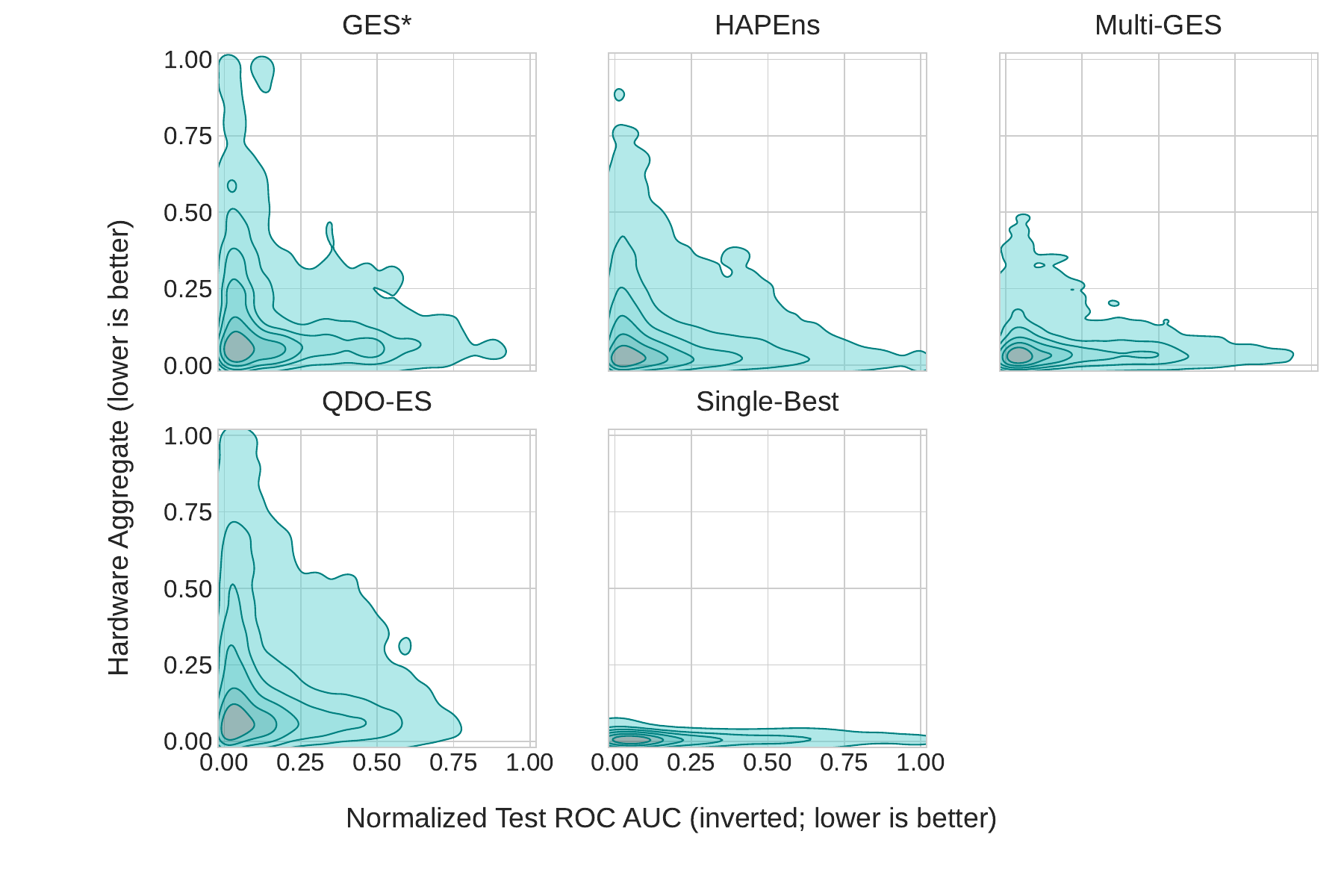}
    \caption{Comparison of constructed ensembles when including cost metrics in the ensembling process. The baselines and the hardware-aware methods in the density plots, produce a clear trend, where the ensembles of the latter methods are more condensed toward the x-axis.}
    \label{fig:density-main}
\end{wrapfigure}

\paragraph{\texttt{EXP2}}  In Figure~\ref{fig:cdp-igp}, the IGD+ results are generally consistent with the HV findings. 
The main difference is the stronger relative performance of Multi-GES, which now significantly outperforms GES and comes close to matching \NAME, to the point that \NAME’s superiority is no longer statistically significant. 
This effect arises because Multi-GES constructs more efficient ensembles, while QDO-ES primarily improves predictive performance (Figure~\ref{fig:density-main}) but at the cost of building more expensive ensembles on average. 
Since IGD+ evaluates solutions with respect to a reference front, Multi-GES benefits disproportionately: a larger share of its efficient solutions lies on the reference Pareto front, reducing the relative advantage of \NAME compared to dominated HV. 
For this reason, we focus on HV in the remainder of the paper, while noting that Multi-GES is particularly strong at exploring the low-cost end of the Pareto front.

\begin{figure}[t]
    \centering
    \begin{subfigure}{0.48\linewidth}
        \centering
        \includegraphics[width=\linewidth]{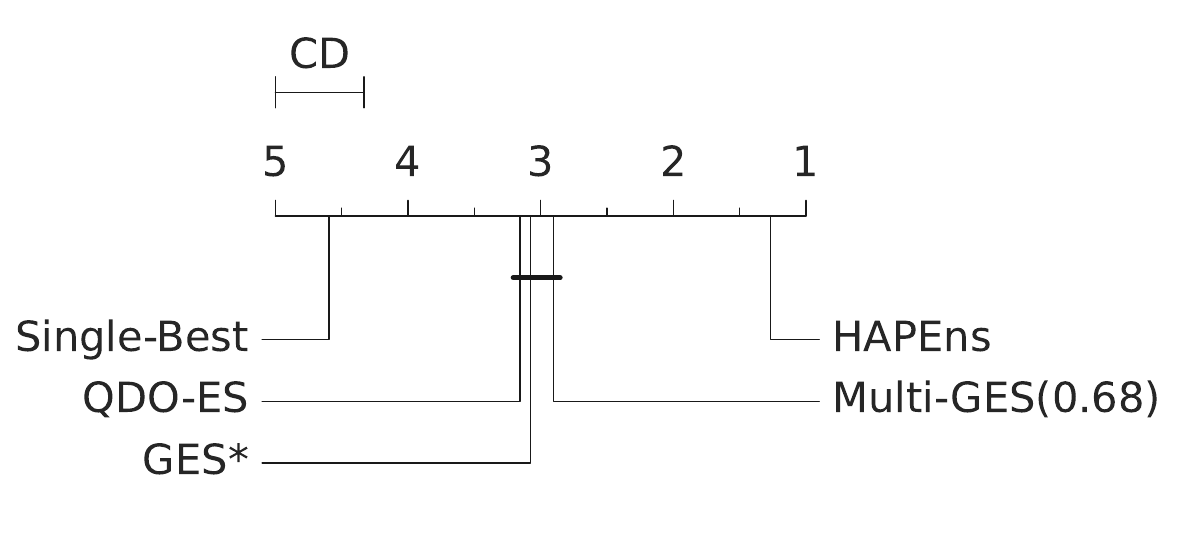}
        \caption{Disk usage.}
        \label{fig:hv_disk}
    \end{subfigure}
    \hfill
    \begin{subfigure}{0.48\linewidth}
        \centering
        \includegraphics[width=\linewidth]{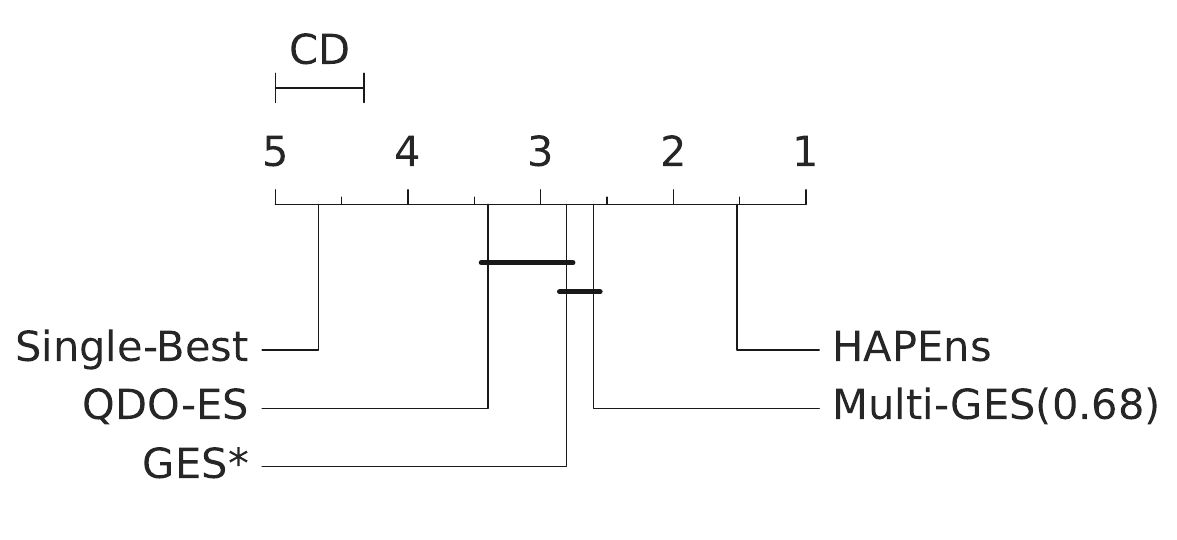}
        \caption{Memory usage.}
        \label{fig:hv_memory}
    \end{subfigure}
    \vskip\baselineskip
    \begin{subfigure}{0.48\linewidth}
        \centering
        \includegraphics[width=\linewidth]{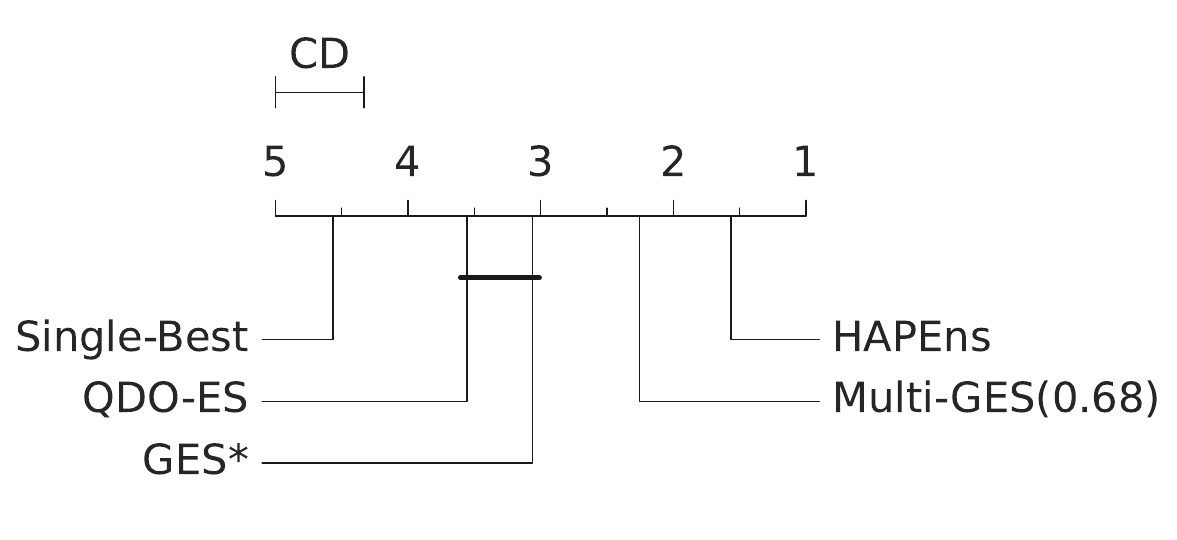}
        \caption{Inference time.}
        \label{fig:hv_time}
    \end{subfigure}
    \caption{Critical difference plots for the hypervolume across different hardware-aware objectives.}
    \label{fig:hv_hardware_metrics}
\end{figure}

\paragraph{\texttt{EXP3}} Looking at the HV results for the individual cost metrics in Figure~\ref{fig:hv_hardware_metrics}, we see in more detail what was already evident in the main results: \NAME\ performs strongly across all metrics. 
The method demonstrates robustness to different hardware considerations, even when the behavior space is defined solely by memory usage. 
Notably, Multi-GES shows a significant improvement over the other baselines when optimizing for inference time. 
This highlights its specialization toward the specific cost metric it uses during ensemble construction. 
It also raises the question of whether incorporating additional cost metrics could lead to further improvements\textemdash but we will leave this for future work.
Since our experiments abstract away from specific hardware configurations, these findings should be viewed as preliminary. 
Overall, these results point to an interesting direction for future research that investigates hardware-aware behavior more directly under diverse configurations and cost measures.

\paragraph{\texttt{EXP4}} Figure~\ref{fig:density-main} shows a density plot of the ensembles constructed by the different methods. 
Compared to Single-Best, all ensemble methods increase hardware costs but also yield clear gains in predictive performance. 
Multi-GES reduces hardware costs relative to GES*, confirming its intended effect. 
QDO-ES and \NAME\ produce similar overall trends, but the ensembles of \NAME\ are more concentrated along the x-axis, indicating lower resource usage. 
These observations clarify and reinforce the improvements of \NAME\ over QDO-ES in terms of hardware efficiency, and likewise of Multi-GES over GES*. 
Overall, the inclusion of cost metrics in the ensemble construction process achieves the desired shift toward more efficient ensembles.  

Figures~\ref{fig:pareto-vs-total} and \ref{fig:total-vs-unique} provide additional insight into the behavior of the tested methods. 
GES* produces 10–15 more ensembles on average than \NAME, yet fewer of them lie on the Pareto front, indicating that many of its ensembles are not useful in this context. 
QDO-ES and \NAME\ both generate a high ratio of unique ensembles, illustrating the effectiveness of the behavior space in promoting diversity. 
By contrast, Multi-GES produces fewer ensembles overall and fewer unique ensembles than GES*, which aligns with the increased difficulty of adding models once hardware costs are incorporated into the selection process.

\subsection*{Ablation (\texttt{EXP5}, \texttt{EXP6})}

\begin{figure}
    \centering
    \begin{minipage}[t]{0.48\linewidth}
        \centering
        \includegraphics[width=\linewidth]{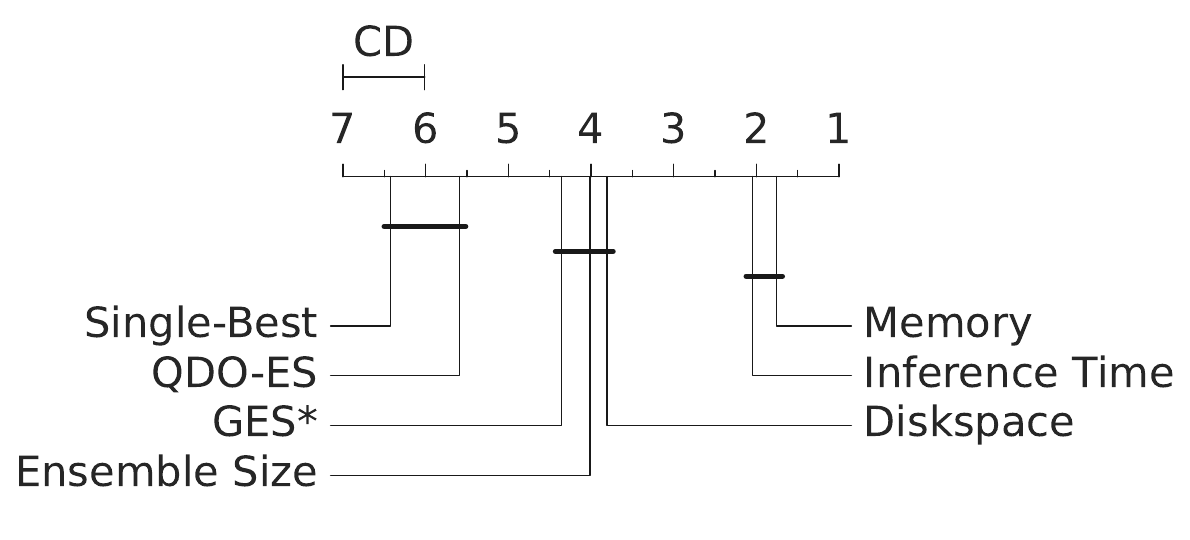}
        \captionsetup{width=\linewidth, format=plain}
        \caption{Comparison of different cost metrics used for \NAME. Memory and inference time perform strongest, but ensemble size is still notable as a proxy cost metric, which does not need additional measurements.}
        \label{fig:variants-comp}
    \end{minipage}
    \hfill
    \begin{minipage}[t]{0.48\linewidth}
        \centering
        \includegraphics[width=\linewidth]{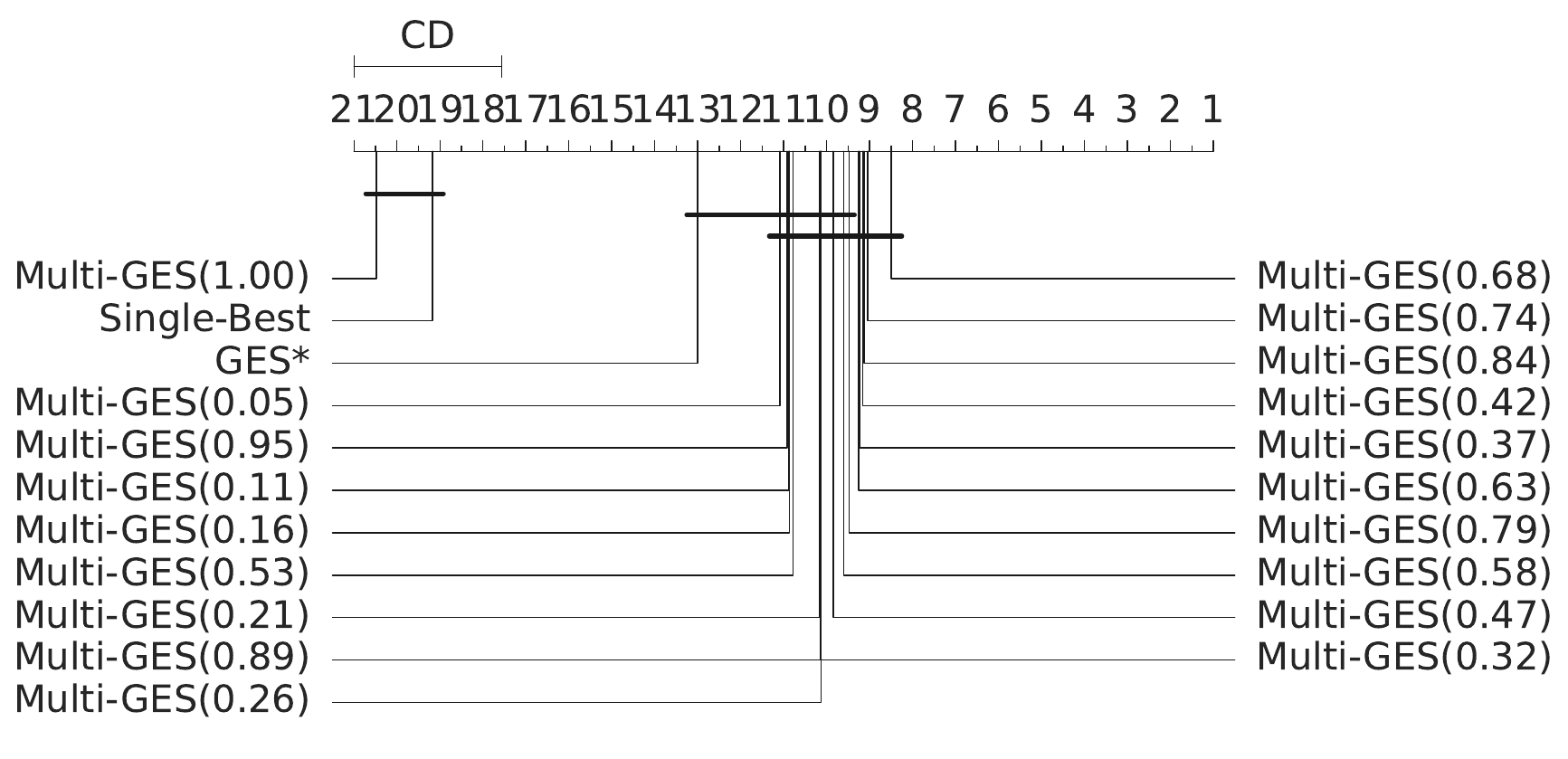}
        \captionsetup{width=\linewidth, format=plain}
        \caption{Comparison of static weights for Multi-GES highlighting the trade-off between predictive performance and hardware costs.}
        \label{fig:mges-comp}
    \end{minipage}
\end{figure}

\paragraph{\texttt{EXP5}} We further evaluated \NAME\ with four different cost metrics: inference time, memory usage, disk usage, and ensemble size. 
The last serves only as a proxy cost metric, yet Figure~\ref{fig:variants-comp} shows that it still provides a competitive signal to balance the trade-off, without requiring additional measurements. 
Among the true cost metrics, memory usage and inference time consistently lead to the strongest results, with memory showing a slight edge. 
These findings highlight that, while the size of the ensemble can act as a lightweight approximation, the use of actual cost metrics yields the most reliable improvements.

\paragraph{\texttt{EXP6}} In Figures~\ref{fig:mges-comp} and \ref{fig:mges-density} we investigate the effect of different static weightings in Multi-GES. 
By gradually increasing the weight on the inference time, the constructed ensembles shift from high-performing but more expensive configurations toward ensembles with lower hardware costs. 
This transition is clearly visible in the density plots, where the mass of ensembles moves closer to the origin of the objective space as the emphasis on inference time increases. 
The trade-off between predictive performance and efficiency becomes apparent: a stronger emphasis on time reduces costs but slightly lowers predictive accuracy, while a weaker emphasis maintains accuracy at the expense of efficiency.
In Figure~\ref{fig:mges-comp} we see a sweet spot, where excessively high or low time weights yield sub-par performance relative to intermediate weightings.
For comparison in the main results, we chose the best performing weight: 0.68.
These results confirm that Multi-GES allows practitioners to explicitly control the desired balance between performance and hardware costs through a weighting mechanism, highlighting its flexibility for different deployment scenarios.

\section{Conclusion}
This work introduced \NAME, a hardware-aware post hoc ensemble selection method that explicitly balances predictive performance and deployment efficiency.
By integrating cost metrics into the ensemble construction process, \NAME\ extends ensemble selection into a multi-objective framework that explores the Pareto front of accuracy and resource usage. 
Across 83 tabular classification datasets, \NAME\ consistently outperforms existing baselines, achieving superior trade-offs under controlled hardware measurement conditions and demonstrating robustness across different cost metrics. 
Ablation studies reveal that memory usage is a particularly effective objective, providing a stable optimization signal and leading to ensembles that generalize well across cost measures. 
Additionally, our experiments show that even simple greedy methods like GES can benefit substantially from static multi-objective weighting, emphasizing the broad potential of hardware-aware ensemble construction. 
To our knowledge, this is the first systematic study of hardware-aware post hoc ensemble selection, opening a new research direction for the AutoML and tabular ML communities. 
Future work may explore dynamic weighting schemes, simultaneous optimization across multiple hardware objectives, task-specific hardware profiling, real-device benchmarking, and integration into end-to-end AutoML pipelines.

\section*{Broader Impact Statement}
This work proposes a general-purpose method for hardware-aware ensemble 
selection in AutoML. The method operates on standard benchmark datasets 
containing no personal or sensitive data. We do not identify negative societal impacts.

\begin{acknowledgements}
L.P. acknowledges funding by the Deutsche Forschungsgemeinschaft (DFG, German Research Foundation) under SFB 1597 (SmallData), grant number 499552394
\end{acknowledgements}


\bibliography{HAPEns}




\appendix

\section{Method}
\begin{algorithm}[ht]
\caption{Crossover and Mutation (adapted from~\cite{purucker_qdo-es_2023})}
\label{alg:crossover_mutation}
\begin{algorithmic}[1]
\STATE \textbf{Input:} Parent repetition vectors $r$, $r'$; model pool $\mathcal{P}$ of size $p$
\STATE Let $S \leftarrow \text{ordered}\bigl(\{i \mid r[i] > 0\} \cup \{i \mid r'[i] > 0\}\bigr)$
\IF{$|S| \geq 3$}
    \STATE Sample cut points $a < b$ uniformly from $\{1, \dots, |S|\}$
    \STATE Initialize $r_{\text{child}} \gets \mathbf{0}$
    \FOR{$k = 1$ \textbf{to} $|S|$}
        \IF{$k \leq a$ \textbf{or} $k > b$}
            \STATE $r_{\text{child}}[S_k] \gets r[S_k]$
        \ELSE
            \STATE $r_{\text{child}}[S_k] \gets r'[S_k]$
        \ENDIF
    \ENDFOR
\ENDIF
\IF{$|S| < 3$ \textbf{or} $r_{\text{child}} = \mathbf{0}$}
    \FOR{$i = 1$ \textbf{to} $p$}
        \STATE $r_{\text{child}}[i] \gets \lceil (r[i] + r'[i]) / 2 \rceil$
    \ENDFOR
\ENDIF
\FOR{attempt $= 1$ \textbf{to} $50$}
    \STATE Sample $j \sim \mathrm{Uniform}(\{1, \dots, p\})$
    \STATE Set $r_{\text{cand}} \gets r_{\text{child}}$ with $r_{\text{cand}}[j] \gets r_{\text{cand}}[j] + 1$
    \IF{$r_{\text{cand}}$ not previously evaluated}
        \STATE Set $r_{\text{child}} \gets r_{\text{cand}}$ and \textbf{break}
    \ENDIF
\ENDFOR
\IF{all 50 attempts failed}
    \STATE Increase increment magnitude or set mutation-after-crossover probability to $1.0$
\ENDIF
\RETURN $r_{\text{child}}$
\end{algorithmic}
\end{algorithm}

\subsection{Enhancements to GES} \label{subsect:enh_ges}
The original GES algorithm optimizes solely for predictive performance without considering other objectives, e.g. hardware costs. This can lead to ensembles that slightly improve accuracy but incur disproportionately higher inference times. To address this, we introduced two enhancements: generating a spectrum of solutions and extending GES to handle multiple objectives.

\paragraph{Spectrum of Solutions}
GES usually outputs a single final ensemble. We modified it to record intermediate ensembles at each iteration, resulting in a set of $n$ ensembles when running for $n$ iterations. Each ensemble adds one model and achieves a higher validation score than the previous one. This creates a spectrum of solutions with varying trade-offs between predictive performance and hardware cost.

\paragraph{Multi-GES}
To explicitly account for hardware efficiency, we extended GES into Multi-GES, introducing a weighted multi-objective scoring system that balances predictive performance and inference time. Both metrics are normalized to ensure scale comparability, and static weights ($\alpha$, $\beta$) control their relative importance.

Algorithm~\ref{algo:multi_ges_normalized} outlines the procedure, where red lines mark the multi-objective components and blue lines indicate the recording of intermediate ensembles.

\begin{algorithm}[htp]
\caption{Multi-GES}
\label{algo:multi_ges_normalized}
\begin{algorithmic}[1]
\STATE Initialize an empty ensemble $E$.
\STATE Initialize the set of candidate models $M$.
\STATE \textcolor{blue}{Initialize an empty list of ensembles $\mathcal{E}$.}
\STATE \textcolor{red}{Set time weight $\beta \in [0, 1]$.}
\STATE \textcolor{red}{Set performance weight $\alpha = 1 - \beta$.}
\WHILE{not finished}
    \STATE Initialize $E'$ as a temporary ensemble.
    \STATE Initialize $J_{\text{best}} \gets \infty$.
    \FOR{each model $m$ in $M$}
        \STATE Form $E' = E \cup \{m\}$.
        \STATE \textcolor{red}{Compute normalized performance $P_{E'}$.}
        \STATE \textcolor{red}{Compute normalized time $T_{E'}$.}
        \STATE \textcolor{red}{Compute objective value: $J_{E'} = \alpha \cdot P_{E'} + \beta \cdot T_{E'}$.}
        \IF{$J_{E'} < J_{\text{best}}$}
            \STATE Update $J_{\text{best}} \gets J_{E'}$.
            \STATE Update $E^* \gets E'$.
        \ENDIF
    \ENDFOR
    \STATE Set $E \gets E^*$.
    \STATE \textcolor{blue}{Append a copy of $E$ to $\mathcal{E}$.}
\ENDWHILE
\RETURN \textcolor{blue}{The set of ensembles $\mathcal{E}$.}
\end{algorithmic}
\end{algorithm}

\section{Experimental Setup} \label{subsect:exp-setup}
\paragraph{Hypervolume (HV):} HV measures the size of the objective space dominated by the solutions on the Pareto front, relative to a chosen reference point. It captures both convergence (how close solutions are to the optimum) and diversity (how well they cover the trade-off surface) in a single scalar value. A larger HV indicates that a method has found better trade-offs between objectives, making it one of the most widely used and Pareto-compliant indicators in multi-objective optimization.

\paragraph{Inverted generational distance plus (IGD+):} IGD+ quantifies how closely the solutions produced by a method approximate a reference Pareto front by measuring the average distance from each point on the reference front to its nearest point in the approximation. It evaluates both how well a method converges to the optimal trade-offs and how evenly its solutions cover the front. Unlike the original IGD, IGD+ only penalizes inferior dimensions, ensuring consistency with Pareto dominance.

\begin{table}[ht]
\centering
\caption{Overview of experiments} \label{tab:exp-overview}
\begin{tabular}{p{2.5cm} p{1.5cm} p{8.5cm}}
\toprule
\textbf{Group} & \textbf{ID} & \textbf{Description} \\
\midrule
Main Results & \texttt{EXP1} & Evaluation of \NAME\ and baseline methods using the HV indicator, calculated from the ROC AUC of the test (inverted to a loss) and the averaged normalized cost metrics: inference time, memory usage, disk space usage. \\
\midrule
\multirow{3}{*}{Details} 
 & \texttt{EXP2} & Comparison of HV and IGD+ values to assess robustness between indicators. \\
 & \texttt{EXP3} & HV evaluation with respect to each individual cost metric (inference time, memory, disk space). \\
 & \texttt{EXP4} & Examining the differences in the construction of the ensemble between \NAME\ and the baseline methods. \\
\midrule
\multirow{2}{*}{Ablation} 
 & \texttt{EXP5} & Analysis of alternative cost metrics in the behavior space of \NAME. \\
 & \texttt{EXP6} & Sensitivity analysis of different weightings in the Multi-GES method. \\
\bottomrule
\end{tabular}
\end{table}

\section{Results}
\begin{figure}
    \centering
    \includegraphics[width=0.8\linewidth]{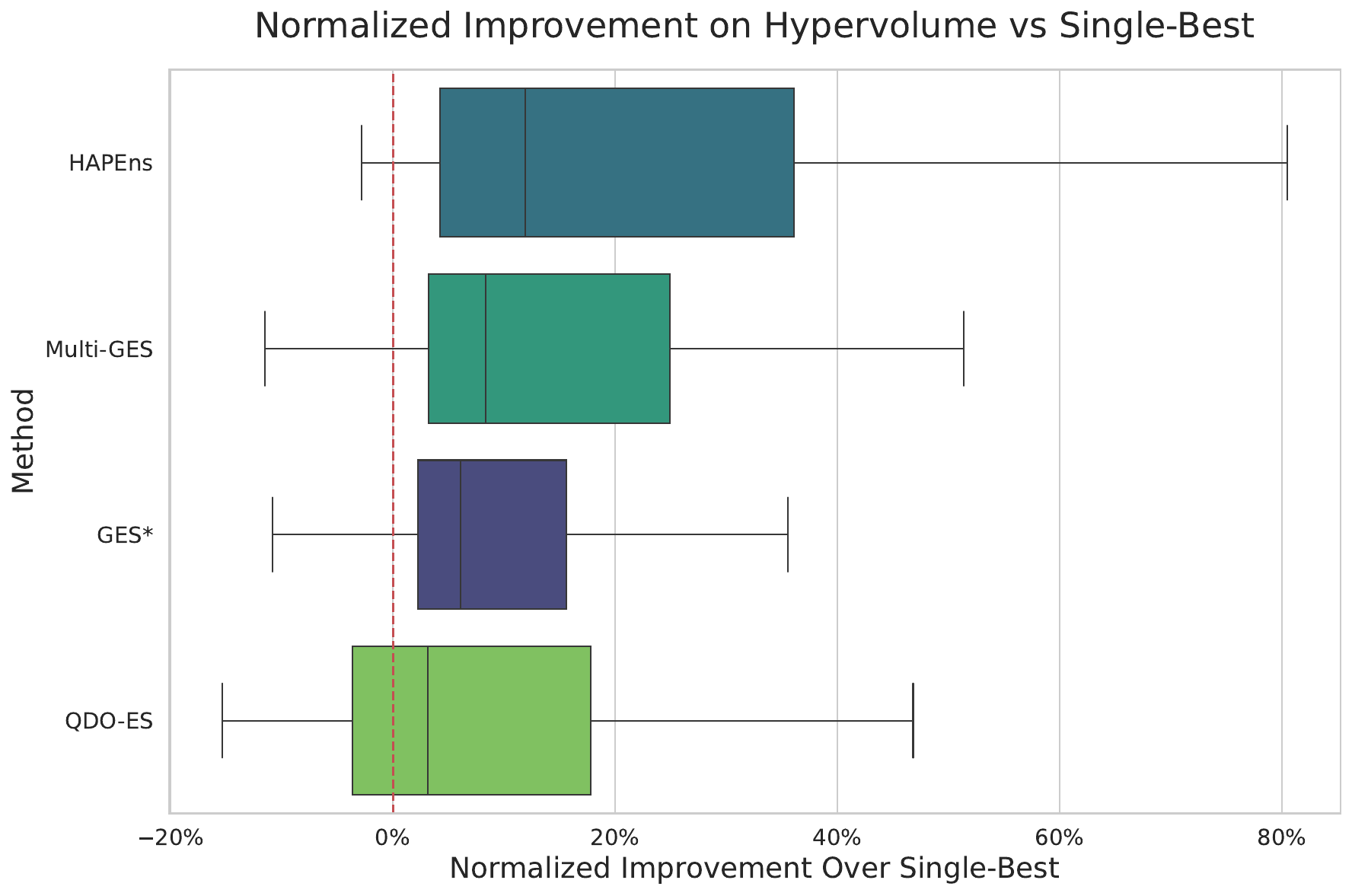}
    \caption{Normalized improvement over the single-best baseline method.}
    \label{fig:main-normalized-improvement}
\end{figure}

\begin{figure}[ht]
    \centering
    \includegraphics[width=0.55\linewidth]{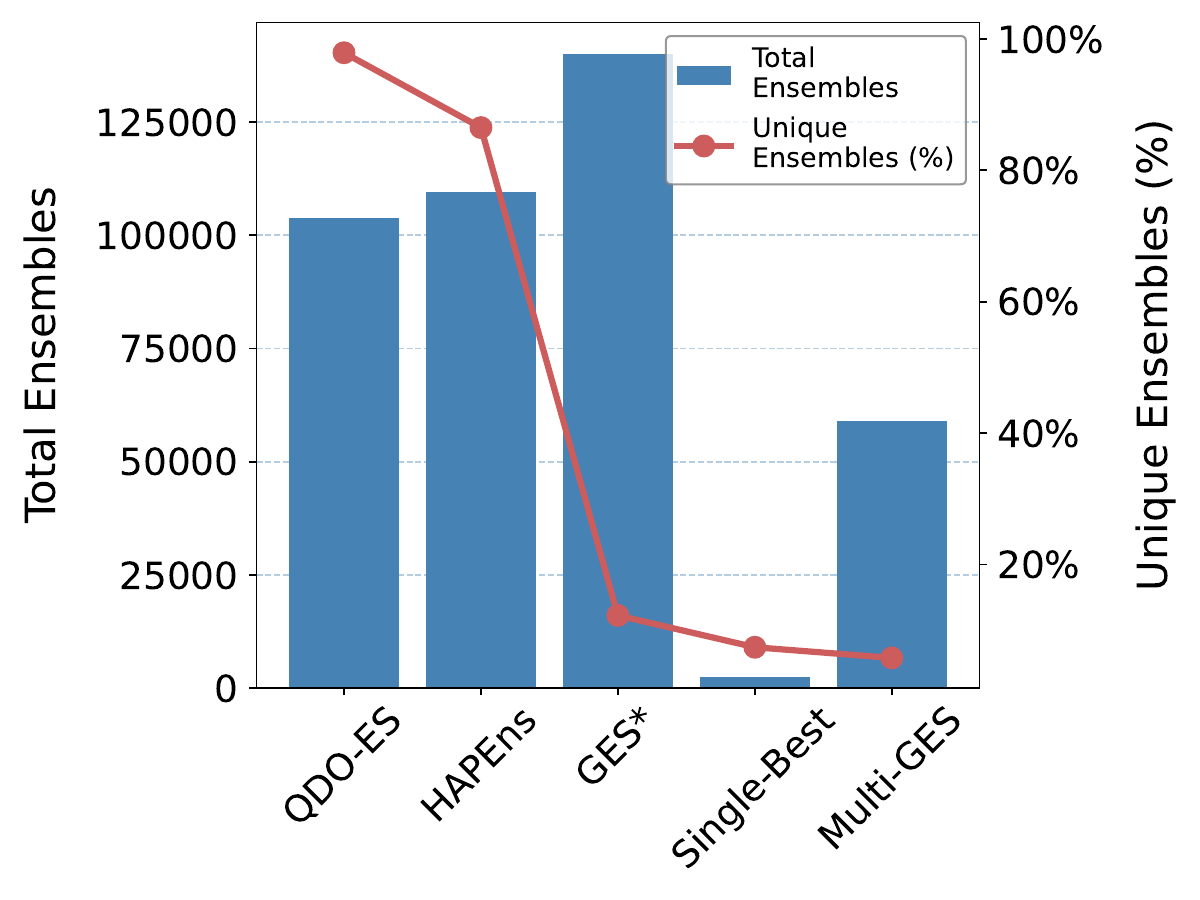}
    \caption{Overall number of ensembles generated per method and how many of these were unique when comparing the specific weight vectors.}
    \label{fig:total-vs-unique}
\end{figure}

\begin{figure}
    \centering
    \includegraphics[width=0.8\linewidth]{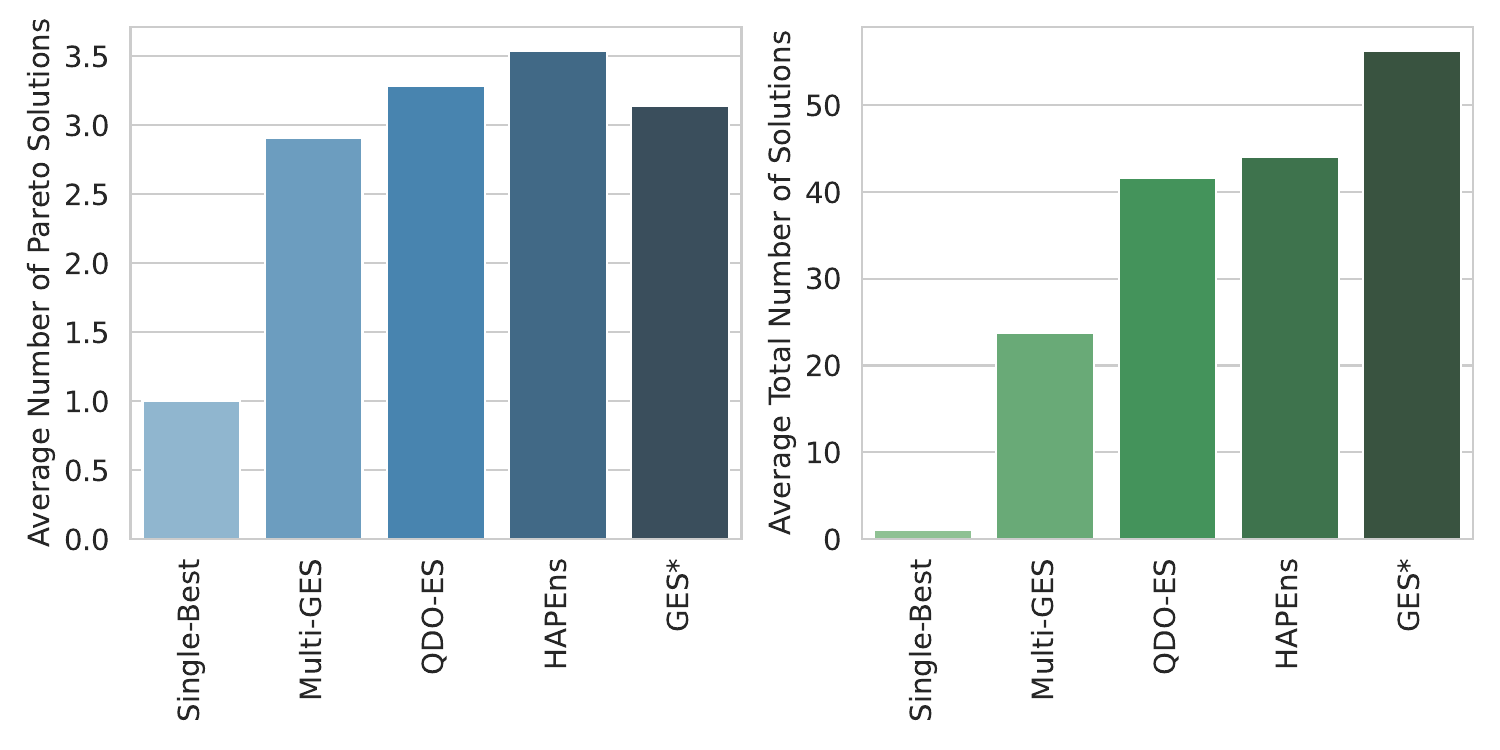}
    \caption{The average number of Pareto ensembles created vs the total number of ensembles created per method on average. The averages are per seed and task to reflect real-world yields of these methods.}
    \label{fig:pareto-vs-total}
\end{figure}

\begin{figure}[p]
    \hspace*{-0.7cm} 
    \includegraphics[width=\textwidth]{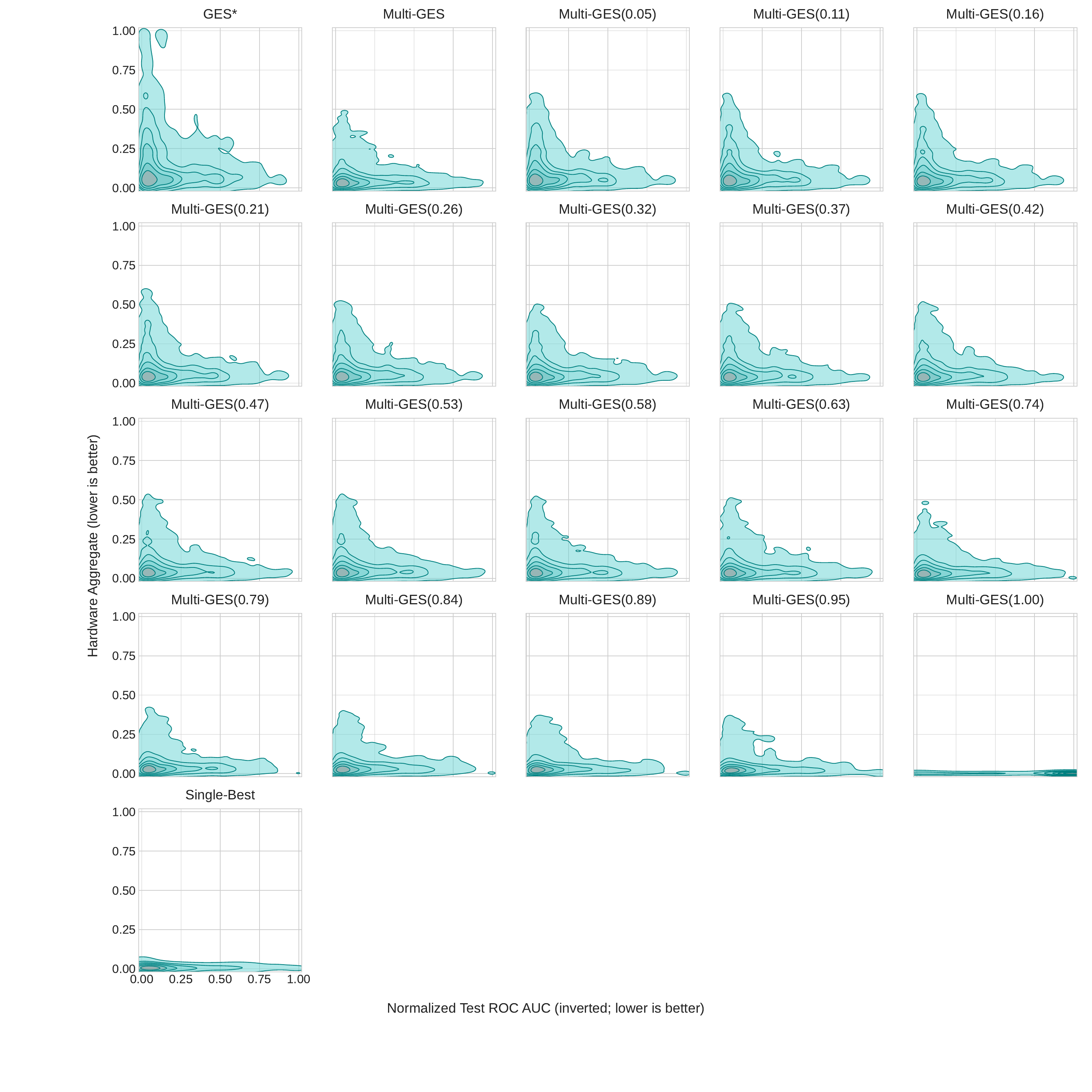}
    \caption{Density of ensembles produced by Multi-GES transitions to less expensive ensembles when increasing the time weight. The sub-plot titled Multi-GES shows the weighting we chose for our experiments: 0.68.}
    \label{fig:mges-density}
\end{figure}

\end{document}